\documentclass[10pt,twocolumn,letterpaper]{article}

\usepackage{cvpr}              
\definecolor{cvprblue}{rgb}{0.21,0.49,0.74}
\usepackage[pagebackref,breaklinks,colorlinks,allcolors=cvprblue]{hyperref}

\usepackage{booktabs}
\usepackage{xcolor}
\usepackage{adjustbox} 
\usepackage{multirow}
\usepackage{makecell}  
\usepackage{booktabs}  

\def\paperID{1408} 
\def\confName{CVPR}
\def\confYear{2026}

\title{TokenPure: Watermark Removal through Tokenized Appearance and Structural Guidance}

\begin{document}

\author{
  Pei Yang\textsuperscript{1} \quad
  Yepeng Liu\textsuperscript{2} \quad
  Kelly Peng\textsuperscript{1} \quad
  Yuan Gao\textsuperscript{1} \quad
  Yiren Song\textsuperscript{1,3 $\dagger$} \\
  \textsuperscript{1}First Intelligence \quad
  \textsuperscript{2}University of California, Santa Barbara \quad
  \textsuperscript{3}National University of Singapore
}

\maketitle

\begingroup
\renewcommand\thefootnote{}
\footnotetext{$\dagger$ Corresponding author.}
\endgroup

\begin{abstract}

In the digital economy era, digital watermarking serves as a critical basis for ownership proof of massive replicable content, including AI-generated and other virtual assets. Designing robust watermarks capable of withstanding various attacks and processing operations is even more paramount. We introduce TokenPure, a novel Diffusion Transformer-based framework designed for effective and consistent watermark removal. TokenPure solves the trade-off between thorough watermark destruction and content consistency by leveraging token-based conditional reconstruction. It reframes the task as conditional generation, entirely bypassing the initial watermark-carrying noise. We achieve this by decomposing the watermarked image into two complementary token sets: visual tokens for texture and structural tokens for geometry. These tokens jointly condition the diffusion process, enabling the framework to synthesize watermark-free images with fine-grained consistency and structural integrity. Comprehensive experiments show that TokenPure achieves state-of-the-art watermark removal and reconstruction fidelity, substantially outperforming existing baselines in both perceptual quality and consistency. Code is released at \href{https://github.com/YoungP2001/TokenPure}{https://github.com/YoungP2001/TokenPure}.
\end{abstract}    
\section{Introduction}
\label{sec:intro}

Currently, the explosive development of generative models\cite{shen2025efficient}, such as Stable Diffusion \cite{rombach2022high} and LLaVA \cite{liu2023visual}, has significantly lowered the barrier to AI creation, leading to a surge in AI-generated content output. Against this backdrop, digital watermarking \cite{liu2024survey,zhao2411sok,liu2025position, he2025distributional} has become increasingly crucial as a key tool for tracking the source of AI content \cite{ci2024ringid,liu2024adaptive,ci2024wmadapter, liu2025context, he2024universally, hui2025autoregressive},  protecting data copyright \cite{liu2025dataset, luo2025imagesentinel, guo2024zeromark, chen2025model, zhang2025leave}, defending against deepfakes \cite{wu2025robust}, and ensuring information transparency.

Robustness lies at the core of watermarking effectiveness, ensuring reliable detection and integrity even in the face of diverse transformations or adversarial attacks \cite{gunn2024undetectable, an2025defending, an2025reinforcement, yang2024gaussian, wen2024tree, song2025idprotector, song2024anti}. However, with the rapid advancement of AI techniques\cite{gong2024neuroclips, zhang2025enhancing, zhang2024mlip, zhang2023mg}, particularly generative models, watermark robustness faces unprecedented challenges in maintaining its reliability and resilience. Therefore, investigating an effective watermark removal method serves as a stress test for current watermarking schemes, revealing their vulnerabilities and guiding the development of future methods with higher robustness standards.

The robustness of existing image watermarking methods varies with the degree of perturbation introduced to either the original image \cite{zhu2018hidden, 2019stegastamp, fernandez2022watermarking, cox2007digital} or the image generation process \cite{wen2024tree, ci2024ringid, fernandez2023stable, ci2024wmadapter}. Low-perturbation watermarks \cite{fernandez2023stable, zhu2018hidden,zhang2019robust}, due to their small perturbation, e.g., small $\ell_2$ distance, of the original content, have relatively limited resistance against naive image manipulation. High-perturbation watermarks \cite{2019stegastamp,saberi2023robustness,zhao2023invisible, ci2024ringid, hui2025autoregressive, ci2024wmadapter, liu2024image} usually introduce substantial modifications to the image in either pixel space or latent space, making it more difficult for adversaries to fully remove the underlying watermark information without destroying the semantics of original images.

Recent advances in generative models have spurred progress in watermark removal methods \cite{zhao2023invisible,an2024benchmarking,liu2024image}. These methods leverage powerful generative models to regenerate the watermarked image, thereby eliminating the embedded watermark signal. These approaches excel at disrupting many low-perturbation watermarks, yet their limitations become evident in handling complex scenarios. For instance, \cite{zhao2023invisible} introduces a regeneration attack that leverages the characteristic noising process of diffusion models, effectively destroying low-perturbation watermarks by iteratively adding noise to the latent representation. However, it performs poorly on high-perturbation watermarks.\cite{an2024benchmarking} seeks to achieve more thorough watermark removal by cascading multiple regeneration iterations. While this strategy improves removal performance to some extent, it also introduces noticeable image degradation and artifacts due to the excessive noising steps. To improve the trade-off between attack performance and image quality, \cite{liu2024image} proposes a controllable regeneration method that reconstructs watermark-free images from clean initial noise using a diffusion model. While this approach substantially enhances removal effectiveness, it also introduces noticeable color and texture deviations in the reconstructed images, compromising visual consistency.

Regenerating a watermarked image from clean noise can effectively eliminate the embedded watermark information, but coarse-grained control often results in noticeable inconsistencies in the reconstructed image. This raises an important question: how can we further enhance reconstruction consistency while still ensuring complete removal of watermark signals?

In this paper, we propose a fine-grained token level regeneration method. We find that heavy compression or high-frequency suppression (e.g., low-token representations, visual prompt encoding \cite{zhai2023sigmoid,radford2021learning}) significantly weakens watermark signals, revealing noise-level embedding vulnerabilities. Additionally, token-level units provide finer-grained reconstruction cues. Motivated by this, we propose TokenPure, a Diffusion Transformer–based conditional framework that removes watermarks without initial watermark-carrying noise. We further compress watermark-agnostic edge information to guide reconstruction via complementary content priors, ensuring thorough removal and structural consistency.
Instead of direct image-to-image translation, TokenPure reformulates the task as conditional generation guided by two complementary token branches: (1) an Appearance Adapter that compresses watermarked images into compact visual tokens to capture color and texture cues; (2) a Layout Controller that extracts edge information and encodes it into structural tokens via VAE to preserve geometry. These branches jointly condition the diffusion transformer (DiT) \cite{dit} to synthesize watermark-free images from visual–structural priors, effectively disentangling watermark signals from the visual content.

Extensive experiments across multiple benchmarks demonstrates that TokenPure achieves superior de-watermarking precision and reconstruction fidelity  It delivers perceptually sharper, more consistent results, moving beyond quantitative metrics.

Our contributions can be summarized as follows: 
\begin{itemize}
\item A novel Diffusion Transformer-based de-watermarking framework, TokenPure, is proposed to enable non-watermark image reconstruction.
\item An Appearance Adapter and a Layout Controller are designed to implement complementary visual-structural conditioning, jointly facilitating accurate and interpretable image purification.

\item Comprehensive quantitative and qualitative evaluations are performed, demonstrating that TokenPure attains state-of-the-art performance in both watermark removal efficacy and reconstruction consistency.
\end{itemize}

\section{Relate works}
\label{sec:related}

\subsection{Image Watermarking Methods}
Image watermarking provides a reliable mechanism for ownership verification and content traceability. 
Early approaches embed watermark information directly into images~\cite{rouhani2018deepsigns, chen2019deepmarks, jia2021entangled}. 
With the rise of large generative models such as Stable Diffusion~\cite{rombach2022high}, watermarking can now be seamlessly integrated during content generation. 
Existing approaches can be broadly divided into \textit{post-hoc} and \textit{in-generation} watermarking. 
Post-hoc methods embed signals after image creation via encoder--decoder architectures~\cite{zhu2018hidden, 2019stegastamp}, optimization~\cite{fernandez2022watermarking}, or transform-based embedding~\cite{cox2007digital}. 
In contrast, in-generation methods \cite{wu2025watermark} modify the generative process itself, such as altering initial noise~\cite{wen2024tree, ci2024ringid,mao2025maxsive}, manipulating latent decoders~\cite{fernandez2023stable, ci2024wmadapter} or semantic embeding ~\cite{arabi2025seal,lee2025semantic}. 
We evaluate our proposed approach under both categories to ensure comprehensive coverage.

\subsection{Image Watermark Removal Methods}
The robustness of watermarks is typically assessed through \textit{removal attacks}~\cite{hu2024transfer, kassis2024unmarker,liang2025watermark, yang2024can}, including editing, regeneration, and adversarial types. 
Editing attacks involve simple manipulations---cropping, compression, rotation, or noise injection---simulating common real-world distortions. 
Most watermarking methods remain stable under such perturbations. 
Recently, \textit{regeneration attacks}~\cite{zhao2023invisible, saberi2023robustness,muller2025black} propose to remove watermarks by re-synthesizing images through the denoising process of pretrained diffusion models or VAE decoding. 
These approaches effectively erase weakly embedded signals but perform poorly against high-perturbation watermarking schemes. 
Alternatively, \textit{adversarial attacks}~\cite{saberi2023robustness, lukas2023leveraging, jiang2023evading, zou2025attention} optimize pixel-level perturbations to deceive detectors, but they require prior knowledge of the watermarking algorithm or model access, making them costly and impractical. 
In this work, we focus on the regeneration-based paradigm and propose a more fine-grained watermark removal strategy to better ensure consistency.
\begin{figure*}[t]
\centering
  \includegraphics[width=\textwidth]{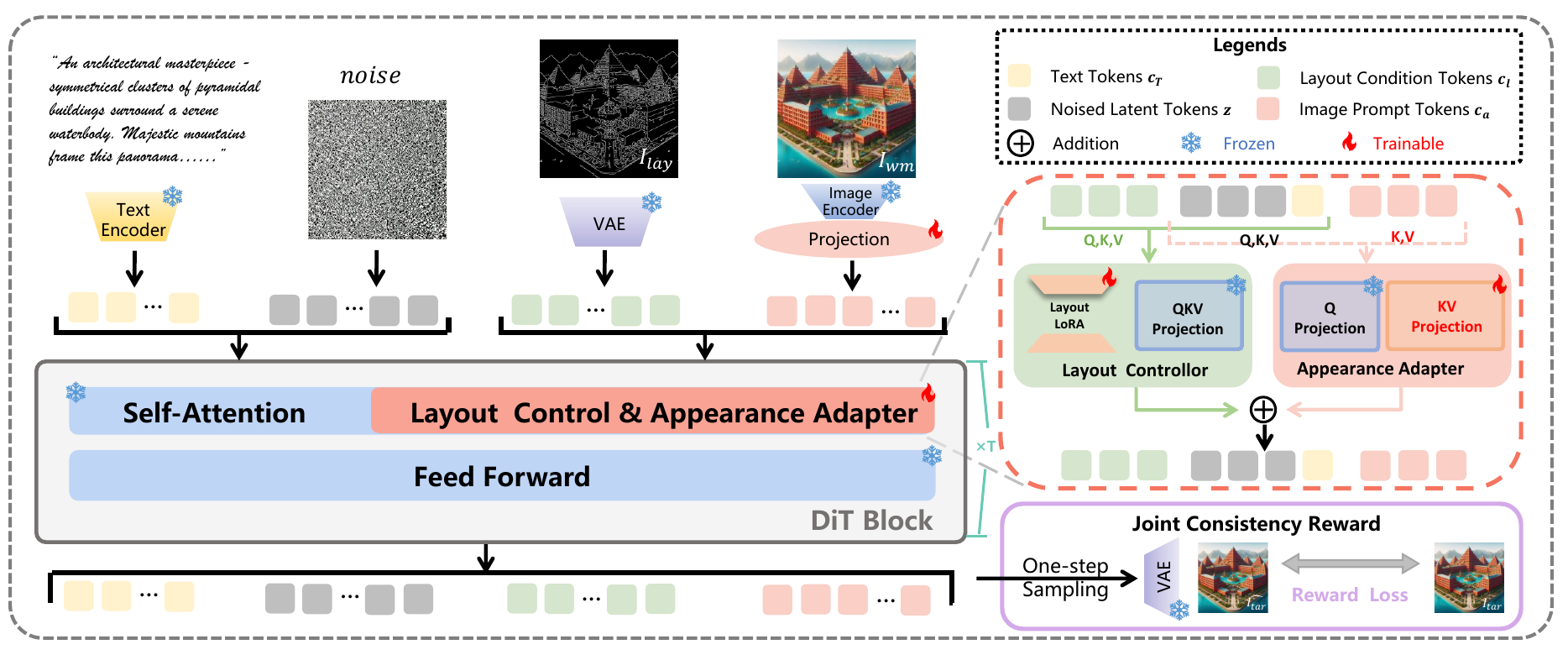}
  \caption{The illustration of TokenPure framwork. It takes text, noise, the layout representation \( I_{\text{lay}} \) (from edge detection and VAE encoding) of the watermarked image \( I_{\text{wm}} \), and image prompt tokens (from the Image Encoder and Projection of \( I_{\text{wm}} \)) as inputs. Within the DiT Block, the Self-Attention and Feed Forward modules process these tokens, while the Layout Control \& Appearance Adapter branch integrates spatial structure and visual details. The Layout Controller uses a LoRA module to optimize layout tokens’ QKV projection, and the Appearance Adapter performs trainable KV projection on image prompt tokens to interact with the Q projection of text and noise tokens via multimodal attention. Finally, the Joint Consistency Reward mechanism refines the result through one-step sampling and pixel-level reward loss, ensuring the reconstructed image’s consistency with the original. } 
  \label{fig:overview}
\end{figure*}

\subsection{Diffusion Models}
Diffusion probabilistic models~\cite{ddim,ddpm} generate data by iteratively denoising Gaussian noise and have achieved remarkable success in image synthesis~\cite{rombach2022high,dit}, editing~\cite{instructpix2pix,p2p,stablemakeup,stablehair}, and 3D generation~\cite{dreamfusion}. 
Stable Diffusion~\cite{rombach2022high} serves as a representative model, employing a UNet backbone to achieve high-quality text-to-image generation and inspired subsequent work~\cite{song2024processpainter, stablehair, stablemakeup, chen2025transanimate}. 
Controllability ~\cite{ma2023magicstick, ma2024followpose, ma2024followyouremoji, ma2025followcreation, ma2025followyourclick, ma2025followyourmotion}  has since improved via auxiliary modules such as ControlNet~\cite{controlnet,li2024controlnet++} and T2I-Adapter~\cite{t2i}, which use multimodal conditions like depth or segmentation. 
Subject-driven generation further evolved from fine-tuning--based personalization~\cite{textualinversion,DB,customdiffusion,lora, zhang2024fast, ssr} to fine-tuning--free approaches~\cite{ye2023ip, ssr}, offering diverse trade-offs between adaptability and efficiency. Recently, the Diffusion Transformer ~\cite{dit} has emerged as a key advancement, replacing the conventional UNet with a highly scalable Transformer architecture operating directly on latent tokens ~\cite{zhang2025easycontrol, lu2025easytext, shi2024fonts, shi2025wordcon, song2025omniconsistency, song2025layertracer, song2025makeanything, wang2025diffdecompose, gong2025relationadapter, guo2025any2anytryon, huang2025arteditor, wan2024grid, jiang2025personalized}. Building upon these advances, we design a controllable diffusion transformer-based framework, enabling precise de-watermarking through token-level conditioning.

\section{Method}
\subsection{Overall Architecture}
\label{sec:overall}

Our framework, \textbf{TokenPure}, is built upon a dual-branch Diffusion Transformer architecture designed to reconstruct watermark-free images from watermarked image tokens without relying on initial noise. As illustrated in Fig.~\ref{fig:overview}, the system consists of two complementary branches: (1) an \textit{Appearance Adapter} Branch that provides visual appearance cues, and (2) a \textit{Layout Controller} Branch that encodes geometric and structural information. Both branches interact through multimodal attention within a unified diffusion backbone based on the Flux Transformer.

\subsection{Appearance Adapter}

Given a watermarked image $I_{wm}$, we first extract its feature sequence using a pretrained image encoder {SigLIP} without applying any noise perturbation. Unlike Flux-IP-Adapter, it only uses the pooled feature as visual prompts, which are more suitable for the classification task. Particularly, we extract the token-level feature of the last hidden layer in SigLIP \cite{zhai2023sigmoid} as the input of our Appearance Adapter, which preserves richer semantic details. The extracted latent features are then projected through a series of linear modules and activators, which map them into a compact set of {visual tokens} $c_a$. These tokens capture color, texture, and fine-scale appearance cues while filtering out high-frequency watermark residuals. The visual tokens are concatenated with the {text tokens} $c_T$ from the frozen Flux model’s T5 \cite{2020t5} text encoder and random noise tokens $z_t$ sampled from a Gaussian distribution, forming a multimodal token sequence. 

A key challenge lies in how to effectively embed semantic features while injecting them without disrupting the prior relationships among the original multimodalities. We have designed a novel Multimodal attention mechanism, which adopts the query $Q$ (including text and noise tokens) from the original DiT module and their frozen Q projection weights ($W_q$). Meanwhile, we create a new key-value ($K_A, V_A$) attention projection module for our visual tokens ($c_a$). The output of the new attention, denoted as $Z_A$, is computed as follows:
\vspace{-7pt} 
\begin{equation}
{Z_A} = \text{Attention}({Q}, {K_A}, {V_A}) = \text{Softmax}\left( \frac{{Q}{K_A}^\top}{\sqrt{d}} \right) {V_A}, 
\end{equation}
\vspace{-1pt} 
where $ {Q} = {(z_t,c_T)}{W}_q, {K_A} = {c}_a{W}_k, {V_A} = {c}_a{W}_v $ are the query, key, and values matrices of the attention operation respectively, and  ${W}_q, {W}_k, {W}_v $ are the corresponding weight projection matrices. $(z_T,c_t)$ represents the concatenation of $z_t$ and $c_T$.

Through a Multimodal Attention mechanism, the Diffusion Transformer integrates the textual condition with the appearance embedding, effectively grounding the reconstruction in clean, semantically aligned visual priors.

\subsection{Layout Controller}
Although semantic branches can achieve semantic consistency between the generated content and the original watermark image through text guidance and high-level semantic features, the inherent Transformer-based sequence modeling characteristic of DiT is prone to problems such as spatial topological structure blurring (such as continuous object edges and distortion of local area geometric relationships) when processing image generation, which is particularly prominent in image reconstruction.

To address this, the Layout Controller branch focuses on the structural aspect of the input, including low-dimensional geometric information such as edge contours and regional boundaries, which can directly act on the token sequence modeling process of DiT. Unlike the conventional control network with redundant parameters and computational overhead, our method integrates the layout condition into the FLUX architecture through a lightweight Low-Rank Adaptation (LoRA) \cite{lora} Module. We first perform edge detection on the watermarked image to extract its layout representation, denoted as $I_{lay}$. This layout is encoded through a Variational Autoencoder (VAE) to obtain a set of \textit{layout tokens} $c_l$ that describe the geometric arrangement of the elements of the scene. The layout tokens are then {concatenated directly} with the denoising latent tokens of the diffusion process. The new QKV projection operations are unmodified as:
\vspace{-2pt} 
\begin{equation}
Q_L, K_L, V_L = Q + \Delta Q_L, K + \Delta K_L, V + \Delta V_L \quad,
\end{equation}
where ${Q, K, V}$ are computed by frozen QKV projection in Flux architecture, and ${\Delta Q_L, \Delta K_L, \Delta V_L}$ are trainable low-rank matrices for each layer. Finally, the output $Z_L$ of Layout Controller is defined as follows: 
\begin{equation}
{Z_L} =\text{Attention}({Q_L}, {K_L}, {V_L}) = \text{Softmax}\left( \frac{{Q_L}{K_L}^\top}{\sqrt{d}} \right) {V_L}.
\end{equation}
\vspace{-2pt}
Ultimately, these clean layout tokens act as fine-grained spatial priors for the transformer. This enables the model to maintain semantic consistency while accurately capturing edge topology and local spatial relationships, thereby effectively mitigating the dispersion of DiT spatial structures during generation.


\subsection{Joint Dual-Branch Reconstruction}

We fuse the Layout Controller attention output ${Z}_L$ with the visual attention output ${Z}_A$ (from the Appearance Adapter) before passing it to the Feed Forward module:
\vspace{-2.2pt} 
\begin{equation}
{Z} = {Z}_L + \alpha \cdot {Z}_A,
\end{equation}
\vspace{-2.2pt} 
where $\alpha $ serves as a tunable scalar parameter that modulates the contribution of the visual prompt to the fused output. During diffusion denoising, the Appearance Adapter guides the model toward accurate local texture restoration, while the Layout Branch enforces global geometric stability. 
By jointly optimizing both branches under shared latent supervision, TokenPure achieves robust watermark removal and superior reconstruction fidelity. 

To further enhance cross-branch coherence, we introduce a \textit{Joint Consistency Reward} mechanism, allowing mutual refinement of structure and visual cues across time steps. During the training process, it adds noise to the unwatermarked image tokens \( z_0 \) and executes the forward diffusion process to obtain the noisy image tokens \( z_t' \). Considered that the flow trajectory used in FLUX.1-dev is rectified flow (ReFlow), 
this is a simple diffusion trajectory: \( z_t = (1 - t)z_0 + t\epsilon \), where \( z_0 \sim p_0 \), \(t \in [0, 1] \) and \( \epsilon \sim {N}(0, 1) \). Since it defines the forward process as a straight path between the data distribution and the noise distribution, the learned velocity field \( v_\theta(z_t, t) \) is used to directly map \( z_t \) to the target distribution during single-step sampling theoretically. The single-step sampling formula can be defined as:
\begin{equation}
z_0' = \frac{z_t - t \cdot v_\theta(z_t, c_T, c_a, c_l, t)}{1 - t}, 
\end{equation}
The core is to modify the linear interpolation path through the velocity field to achieve single-step denoising from the noisy state \( z_t \) to the generated tokens \( z_0' \).)
Particularly, When the added noise \( \epsilon \) is small, the target image tokens \( z_0 \) can be similar to \( z_0' \).

Then, we directly decode cleaned tokens \( z_0' \) into dewatermarked image \( \hat {I_{tar}} \) for reward fine-tuning and calculate the reward consistency loss in the pixel space:  
\begin{equation}
{L}_{{re}} = {L}( {I_{tar}}, \hat {I_{tar}})= \frac{1}{H \times W} \sum_{i=1}^{H} \sum_{j=1}^{W} \left( z_{0(i,j)} - z_{0(i,j)}' \right)^2,
\end{equation}
where $H$ and $W$ represent the height and width of the image ${I_{tar}}$ and $\hat {I_{tar}} $ respectively, and $H \times W$ denotes the total number of pixels. In the reward fine-tuning stage, the pre-trained DiT model are frozen, and only dual branch networks are updated. This ensures that the generation capability of the model is not affected.

Finally, the loss is the combination of diffusion training loss ${L}_{{diff}}$ and the reward loss:
\begin{equation}
{L}_{{total}} = 
\begin{cases} 
{L}_{{diff}} + \lambda \cdot {L}_{{re}}, & {if } t \leq t_{{thre}}, \\
{L}_{{diff}}, & {otherwise},
\end{cases}
\end{equation}
where \( t_{{thre}} \) denotes the timestep threshold and $\lambda$ controls the strength of reward. The diffusion loss \({L}_{{diff}}\) enforces consistency in data distribution, while the reward loss \({L}_{{re}}\) imposes constraints on pixel-level consistency. These two losses complement each other, enabling the model to purify latent representations while maintaining consistent spatial alignment, leading to state-of-the-art (SOTA) performance in both de-watermarking accuracy and visual consistency.

\begin{figure*}[h]
\centering
  \includegraphics[width=\textwidth]{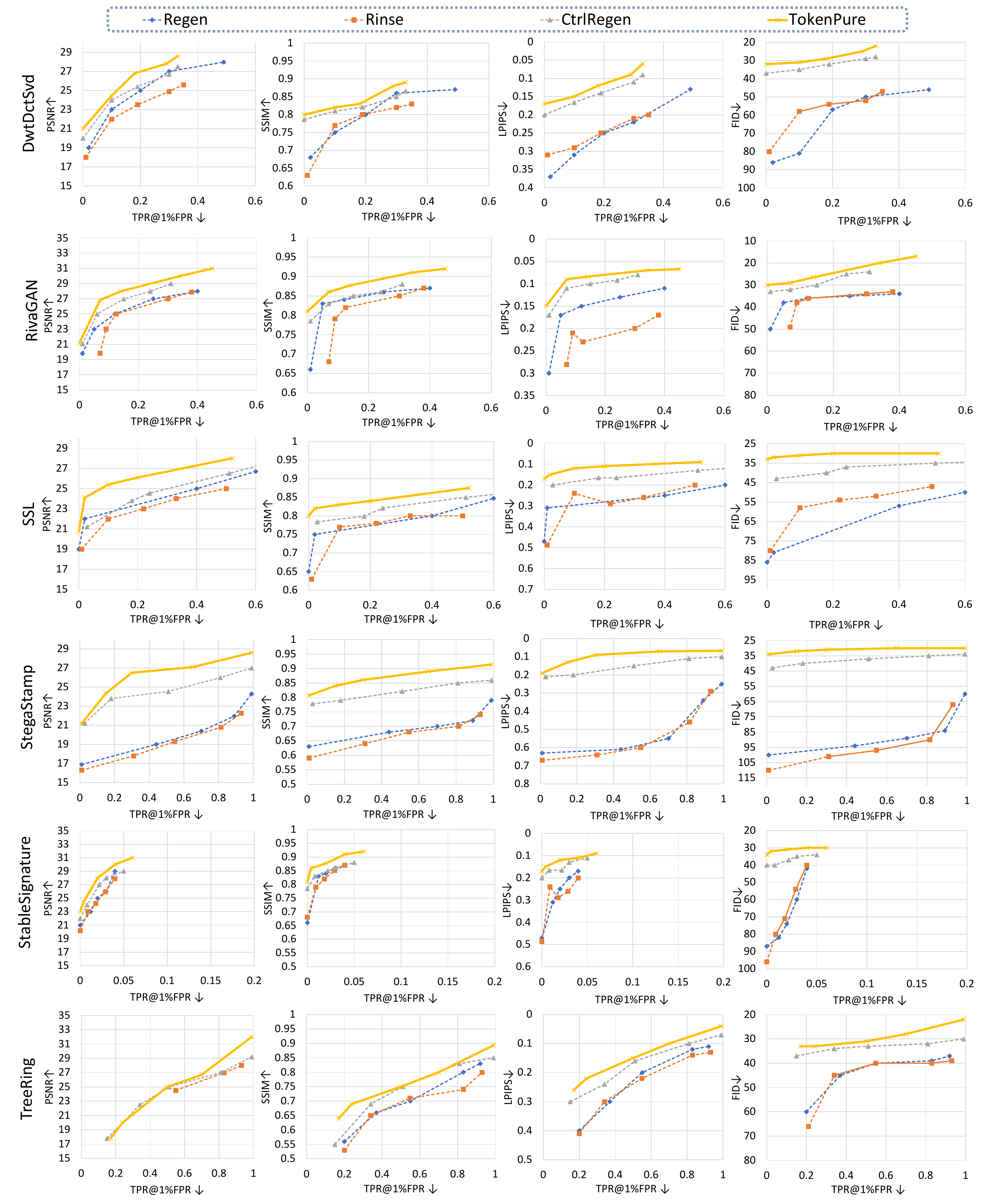}
  \caption{Performance of TokenPure compared to other methods with a varying number of noise strengths on different watermarks, respectively. We invert the LPIPS/FID scores to ensure that the top-left represents better performance across all figures. }
  \label{fig:line_chart}
\end{figure*}

\section{Experiment}

\subsection{Experimental Setting}
\noindent \textbf{Experimental Setup} Our model is initialized with the frozen FLUX.1-dev \cite{flux2024} (DiT architecture), trained on 4 H20 GPUs for 100,000+ iterations with an accumulated batch size of 4. We use the AdamW optimizer \cite{loshchilov2017decoupled} with bfloat16 mixed-precision training and an initial learning rate of $1 \times 10^{-4}$. The Appearance Adapter Module employs SigLIP as the visual prompt encoder, with compressed features output via a two-layer linear projection network and GELU activation \cite{hendrycks2016gaussian}. The Layout Controller Module uses LoRA (rank=128) for efficient fine-tuning, preserving pre-trained generation capabilities. All inference is conducted on a single H20 GPU.

\noindent \textbf{Dataset and Benchmark} For the joint training of the Appearance Adapter and Layout Controller, we use 10 million image-text pairs from LLaVA-NeXT-Data, Dalle3 1 Million+ High Quality Captions \cite{Egan_Dalle3_1_Million_2024}, and Conceptual 12M (CC12M) \cite{changpinyo2021cc12m}. For testing watermark attack performance, we construct a test dataset with 1000 images (synthetic/real from Echo-4o-Image \cite{ye2025echo} and ImageNet \cite{deng2009imagenet}) and 1000 GPT-4o-generated prompts, used to generate watermarked images via six methods: post-hoc (DwtDctSvd \cite{cox2007digital}, RivaGAN \cite{zhang2019robust}, SSL \cite{fernandez2022watermarking}, StegaStamp \cite{2019stegastamp}) and in-generation (TreeRing \cite{wen2024tree}, StableSignature \cite{fernandez2023stable}).

\noindent  \textbf{Baseline Methods} 
We compare with three regeneration-based baselines: Regen \cite{zhao2023invisible} (100-step noise injection), Rinse-2xDiff \cite{an2024benchmarking}, and CtrlRegen \cite{liu2024image},with Stable Diffusion-v1.5 as the base model.

\noindent  \textbf{Evaluation Metrics} The efficacy of watermark removal is evaluated using two core metrics: average Bit Accuracy (BitAcc) and TPR@1\%FPR (True Positive Rate at 1\% False Positive Rate). BitAcc quantifies correctly recovered bits in multi-bit watermarking, with lower values meaning more thorough watermark destruction. TPR@1\%FPR assesses removal stealth by calculating true positive rate under 1\% false positive rate, where smaller values indicate harder-to-detect effects. For clarity, both metrics are reported for watermarked images before (BitAcc B, TPR@1\%FPR B) and after (BitAcc A, TPR@1\%FPR A) attack.

In terms of image quality assessment, metrics include Peak Signal-to-Noise Ratio (PSNR) and Structural Similarity Index (SSIM), which evaluate at the pixel level. Additionally, we also adopt Learned Perceptual Image Patch Similarity (LPIPS) \cite{zhang2018unreasonable}, Deep Image Structure and Texture Similarity (DISTS) \cite{ding2020image}, Quality-Align (Q-Align) \citep{wu2023qalign}, CLIP-based Image Quality Assessment (CLIP-IQA) \cite{wang2022exploring}, and Fréchet Inception Distance (FID) \cite{heusel2017gans} as evaluation metrics, which can measure the similarity and quality differences between images from the perspective of human visual perception.

\subsection{Comparison and Evaluation}

\noindent \textbf{Quantitative Evaluation} Figure \ref{fig:line_chart} presents a comparative analysis of TokenPure against three alternative methods under varying noise intensities, evaluated across four low-perturbation watermarking schemes (DwtDctSvd, RivaGAN, SSL, and StableSignature) and two high-perturbation counterparts (StegaStamp and TreeRing). For low-perturbation watermarks, the number of noising steps was set to $\{50, 100, 150, 200, 300\}$. Compared to Regen and Rinse, TokenPure not only maintains superior image quality and similarity but also achieves state-of-the-art watermark removal performance. Specifically, under the same TPR@1\%FPR criterion, our method outperforms competing approaches in both pixel-level metrics (PSNR/SSIM) and perceptual quality assessments (Q-Align/FID). 

For high-perturbation watermarks, the noising steps were extended to $\{100, 200, 300, 400, 500\}$. While Regen and Rinse can achieve watermark removal with excessive noise injection, they cause substantial degradation to the original image content. Notably, when the number of noising steps exceeds 500, the regenerated images lose critical original information, rendering the watermark removal process meaningless. Although CtrlRegen can disrupt high-perturbation watermarks while preserving basic image content, it suffers from significant deviations in semantic information.
In contrast, TokenPure achieves effective watermark removal while retaining the integrity of the original image information, striking a better balance between watermark erasure efficacy and image consistency. Additional quantitative results across other evaluation metrics are provided in the supplementary materials.


\begin{figure*}[t]
\centering
  \includegraphics[width=\textwidth]{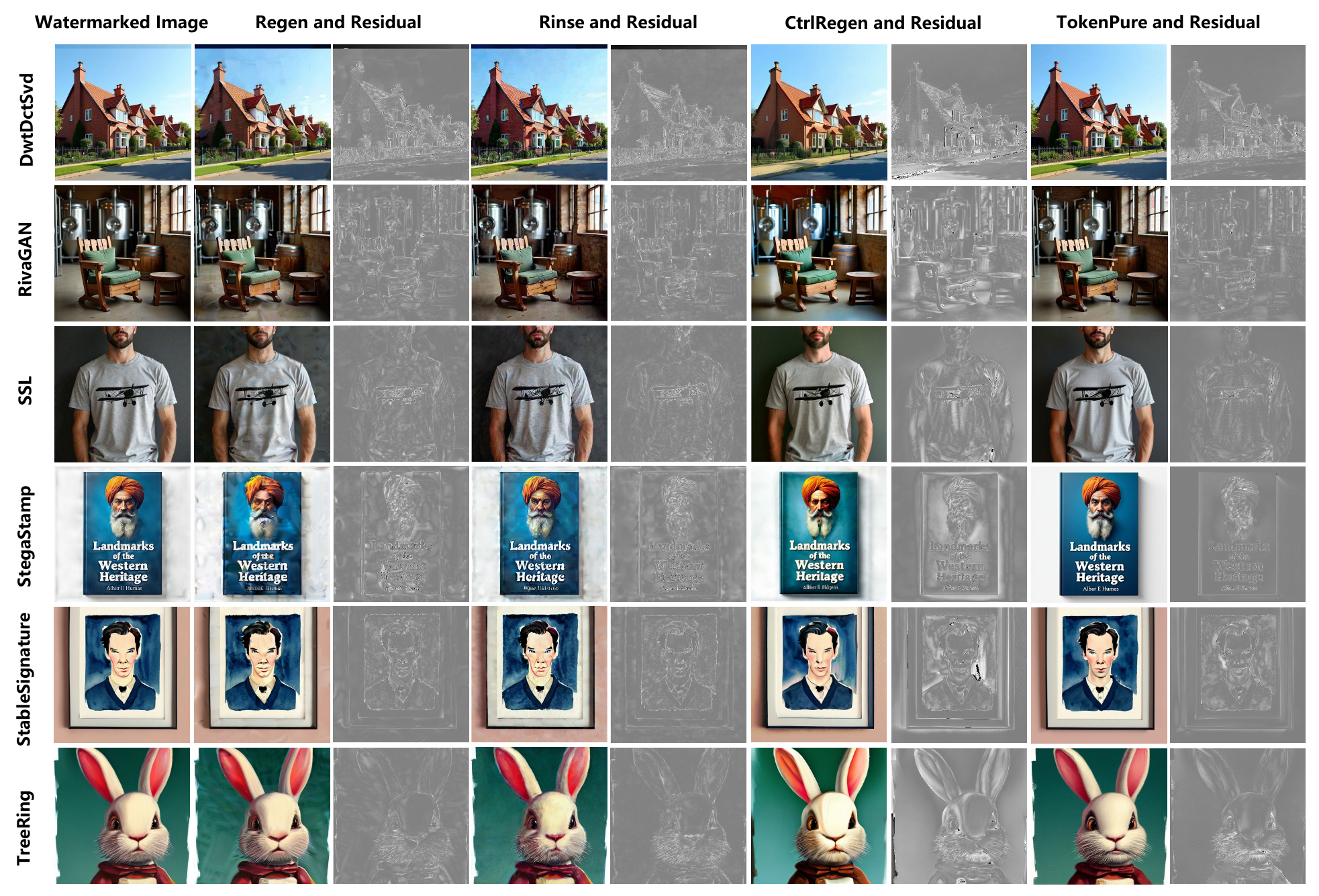}
  \caption{Qualitative comparison of different watermark removal attacks on different watermarking methods. It shows that our TokenPure preserves high visual consistency and quality under various watermarks. } 
  \label{fig:visual_com}
\end{figure*}

\begin{table}[b]
\centering
\caption{Performance comparison of visual token compression for Watermark removal. Results show that visual tokens achieve superior watermark weakening capability.}
\adjustbox{max width=\linewidth}{
\begin{tabular}{
  c |                          
  >{\centering\arraybackslash}m{3cm} |  
  c                           
  c                           
  >{\centering\arraybackslash}m{1.4cm}   
  >{\centering\arraybackslash}m{1.4cm}    
}
\hline
Watermarks & method & BitAcc B $\uparrow$ & BitAcc A $\downarrow$ 
& TPR@1\% FPR B $\uparrow$  
& TPR@1\% FPR A $\downarrow$ \\  
\hline
\multirow{3}{*}{StegaStamp} & Resize image control & 1.0 & 0.6584 & 1.0 & 0.5420 \\
& semantic token & 1.0 & 0.4821 & 1.0 & 0.0025 \\
& semantic token \& canny control & 1.0 & 0.4947 & 1.0 & 0.0080 \\  
\hline
\multirow{3}{*}{TreeRing} & Resize image control & 1.0 & 0.8268 & 1.0 & 0.4087 \\
& semantic token & 1.0 & 0.7524 & 1.0 & 0.1270 \\
& semantic token \& canny control & 1.0 & 0.7670 & 1.0 & 0.1580 \\  
\hline
\end{tabular}
}
\label{tab:model_Ablation1}
\end{table}
\vspace{-1pt}

\noindent \textbf{Qualitative Evaluation} Figure \ref{fig:visual_com} shows visual comparisons of watermark removal attacks across methods, with each watermarking column including an attacked image and a residual map (difference between attacked and original unwatermarked images).
In low-perturbation scenarios, TokenPure’s generated images are highly consistent with originals in texture and color. For RivaGAN, furniture details in TokenPure’s result fully match the original, while Regen, Rinse, and CtrlRegen show obvious local deviations in color and texture. 
For high-perturbation watermarks, TokenPure removes watermarks while preserving original core features. Regen and Rinse leave background artifacts due to watermark noise. Though CtrlRegen avoids noise influence for a clean background, it has slight style deviations. In contrast, TokenPure’s images are nearly identical to originals in color saturation. 

Furthermore, we conduct a dedicated study on inference efficiency (detailed in the supplementary materials) to provide a practical efficiency-performance trade-off. Additionally, we supplement more visual results for related experiments to enhance result interpretability.

\begin{table*}[htpb]
  \centering
  \footnotesize  
  \caption{Ablation study on the effectiveness of the Layout Controller (LC), Appearance Adapter (AA), and Joint Consistency Reward (JCR) mechanism in image reconstruction. The best results are denoted as \textbf{bold}. The results reveal that AA and LC considerably improve the image quality, while the JCR mechanism further enhances image consistency.}
  \renewcommand{\arraystretch}{0.85}  
  \begin{tabular*}{\linewidth}{@{\extracolsep{\fill}}c|c|c|c|c|ccccccc@{}}
    \toprule
    \textbf{Model} & \textbf{AA} & \textbf{Redux} & \textbf{LC} & \textbf{JCR} & {PSNR (dB) $\uparrow$} & {SSIM $\uparrow$} & {LPIPS $\downarrow$} & {DISTS $\downarrow$} & {Q-Align $\uparrow$} & {CLIP-IQA $\uparrow$} & {FID $\downarrow$} \\
    \midrule
    \textbf{A} & $\checkmark$ &  &  &  & 15.89 & 0.6354 & 0.4483 & 0.2757 & 3.609 & 0.7379 & 79.59 \\
    \textbf{B} &  & $\checkmark$ &  &  & 13.85 & 0.5858 & 0.5053 & 0.2955 & 3.629 & 0.7249 & 85.58 \\
    \textbf{C} &  & $\checkmark$ & $\checkmark$ &  & 17.44 & 0.6713 & 0.2986 & 0.2004 & 4.012 & 0.7641 & 56.85 \\
    \textbf{D} & $\checkmark$ &  & $\checkmark$ &  & 20.29 & 0.7937 & 0.2088 & 0.1395 & 4.092 & 0.7689 & 35.30 \\
    \textbf{E} & $\checkmark$ &  & $\checkmark$ & $\checkmark$ & \textbf{21.19} & \textbf{0.8074} & \textbf{0.1920} & \textbf{0.1347} & \textbf{4.197} & \textbf{0.7743} & \textbf{33.34} \\
    \bottomrule
  \end{tabular*}
  \label{tab:model_Ablation2}
\end{table*}

\begin{table}[h]
\centering
\small
\caption{Evaluation of watermarking methods against VAE reconstruction. It is proven that different watermarks exhibit strong robustness to VAE.}
\adjustbox{max width=\linewidth}{%
\begin{tabular}{c|cccc}
\hline
Watermarks & BitAcc B $\uparrow$ & BitAcc A $\downarrow$ & TPR@1\%FPR B $\uparrow$ & TPR@1\%FPR A $\downarrow$ \\ \hline
SSL        & 1.0      & 0.8872                & 1.0          & 0.8250                     \\
StegaStamp & 1.0      & 0.9991                & 1.0          & 1.000                      \\
TreeRing   & 1.0      & 1.000                 & 1.0          & 1.000                      \\ \hline
\end{tabular}
}
\label{tab:vae_rec}
\end{table}

\subsection{Model Analysis}
In this section, we conduct an ablation study on the effectiveness of our proposed method. 

\textbf{Ablation study} 
We compared three pipelines in watermark removal, including directly inputting the original image as a visual prompt, using the semantic token prompt, and combining semantic tokens with edge control. As shown in Table \ref{tab:model_Ablation1}, in terms of both bit accuracy destruction and the stealth of watermark removal, the token compression-based methods significantly outperform simple image resizing control. This indicates that the semantic guidance strategy based on token compression can effectively destroy the bit structure of watermarks and reduce their recognizability. Particularly, the experimental results on two types of watermarks with strong perturbations further verify the universal effectiveness of token compression in watermark removal tasks.

We also conduct an ablation study to verify the effectiveness of TokenPure’s modules (see in Table \ref{tab:model_Ablation2}), confirming each module’s necessity and synergistic effect. To validate the Appearance Adapter (AA), we use FLUX.1 Redux, a prompt-based reconstruction adapter, as a baseline.

Through comparison, it is found that the core metrics of Model A/D (with AA) are comprehensively superior to those of Model B/C (with Redux). This indicates that when processing visual tokens, AA can more accurately preserve details such as color and texture, whereas Redux tends to focus on style transfer with weaker guidance from semantic information. For the Layout Controller (LC), horizontal comparison between Model A/B (without LC) and Model C/D (with LC) shows that with the same semantic adapter, all metrics are significantly optimized after adding LC. This proves that LC can effectively address the problem of spatial layout blurriness and enhance the structural integrity of images. Furthermore, comparing Model D and E reveals that all metrics are improved after integrating the Joint Consistency Reward (JCR), especially indicators like PSNR. This demonstrates that the pixel-level consistency constraint of JCR can further reduce reconstruction errors on the basis of the dual-branch architecture, enhancing the visual consistency between the reconstructed image and the original image, and ultimately achieving state-of-the-art performance.

To further demonstrate our method’s effectiveness, we tested VAE-based watermark attacks. Table \ref{tab:vae_rec} shows that after VAE reconstruction, all three watermarking methods retain high bit accuracy and detectability (close to 1), indicating strong robustness to VAE reconstruction and that VAE alone cannot effectively destroy or remove watermarks. We also provide a more efficient version and additional visual comparison results in the supplementary materials.
\begin{figure}[t]
\centering
  \includegraphics[width=0.5\textwidth]{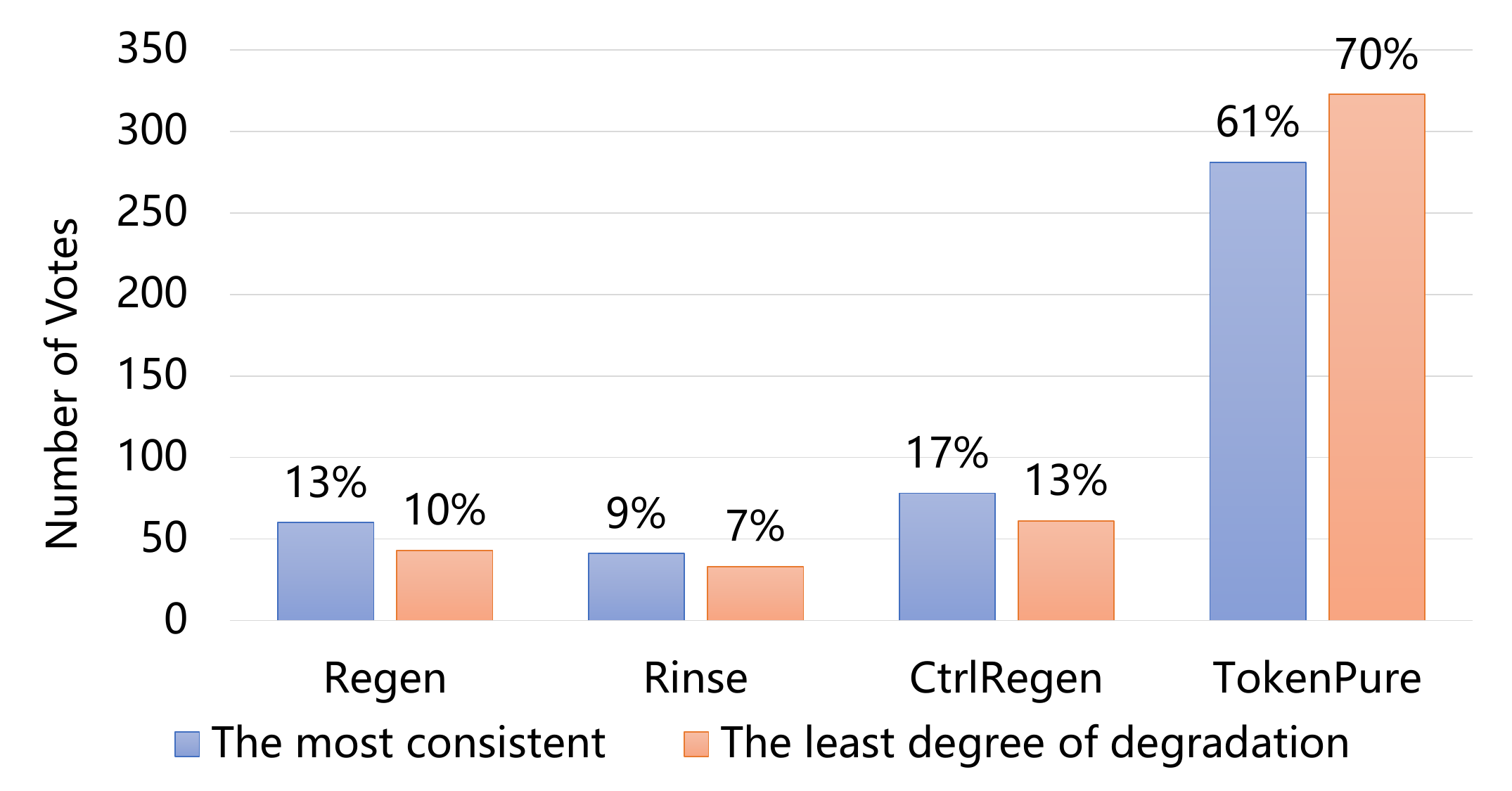}
  \caption{User study results. Our TokenPure is preferred by human voters over other baselines. The results show the advantages of Tokenpure in both image consistency and degradation control.} 
  \label{fig:user_study}
\end{figure}

\subsection{User Study}
For comprehensive watermark removal evaluation, we conducted a user study comparing TokenPure with baselines (Regen, Rinse, CtrlRegen). 46 voters assessed 10 image groups (each with 4 removal results and 1 unwatermarked reference), answering two questions: 1) select the result most consistent with the reference, and 2) select the one with the least degradation/artifacts. As shown in Figure \ref{fig:user_study}, TokenPure led the 920 total votes (46×10×2), winning 61\% for “most consistent” and 70\% for “least degradation”—far exceeding others. This confirms its superiority in both reference consistency and perceptual quality.

\section{Conclusion}
In this paper, we introduced TokenPure, a novel dual-branch Diffusion Transformer framework specifically engineered for the  watermark removal task. By integrating an Appearance Adapter and a Layout Controller, it effectively addresses the critical challenge of removing watermarks while safeguarding both the quality and structural consistency of the original image. Extensive experimental results demonstrate that TokenPure achieves state-of-the-art performance across comprehensive benchmarks, successfully overcoming the limitations of traditional methods in handling high-perturbation watermarks, which solidifies its standing as a robust solution for watermark removal research.

{
    \small
    \bibliographystyle{ieeenat_fullname}
    \bibliography{main}
}

\clearpage
\setcounter{page}{1}
\maketitlesupplementary

\noindent  \textbf{Details about Evaluation Metrics} In terms of image quality assessment, metrics are categorized into two groups based on evaluation perspectives:
Pixel-level metrics, including Peak Signal-to-Noise Ratio (PSNR) and Structural Similarity Index (SSIM). Peak Signal-to-Noise Ratio (PSNR) quantifies distortion mathematically by comparing signal strength to noise, with higher values indicating less distortion. Structural Similarity Index (SSIM) focuses on structural consistency across brightness, contrast, and spatial structure, with values closer to 1 reflecting stronger similarity to the original image at the pixel level.

Perceptual-level metrics, encompassing Learned Perceptual Image Patch Similarity (LPIPS), Deep Image Structure and Texture Similarity (DISTS), Quality-Align (Q-Align), CLIP-based Image Quality Assessment (CLIP-IQA), and Fréchet Inception Distance (FID). Both LPIPS and DISTS employ deep neural networks to quantify perceptual differences; consistently, lower metric values correspond to higher perceived similarity. DISTS is specifically designed to heighten sensitivity to structural and textural integrity, granting it greater tolerance for subtle color shifts. Q-Align integrates text-image alignment and visual quality, with higher scores denoting better alignment with textual instructions. CLIP-IQA, relying on cross-modal models, assesses quality or alignment without reference images, where higher values signal superior perceptual quality. FID measures the distance between feature distributions of generated and real images, with lower values indicating greater realism and diversity in generated content.

\begin{figure*}[h]
\centering
  \includegraphics[width=\textwidth]{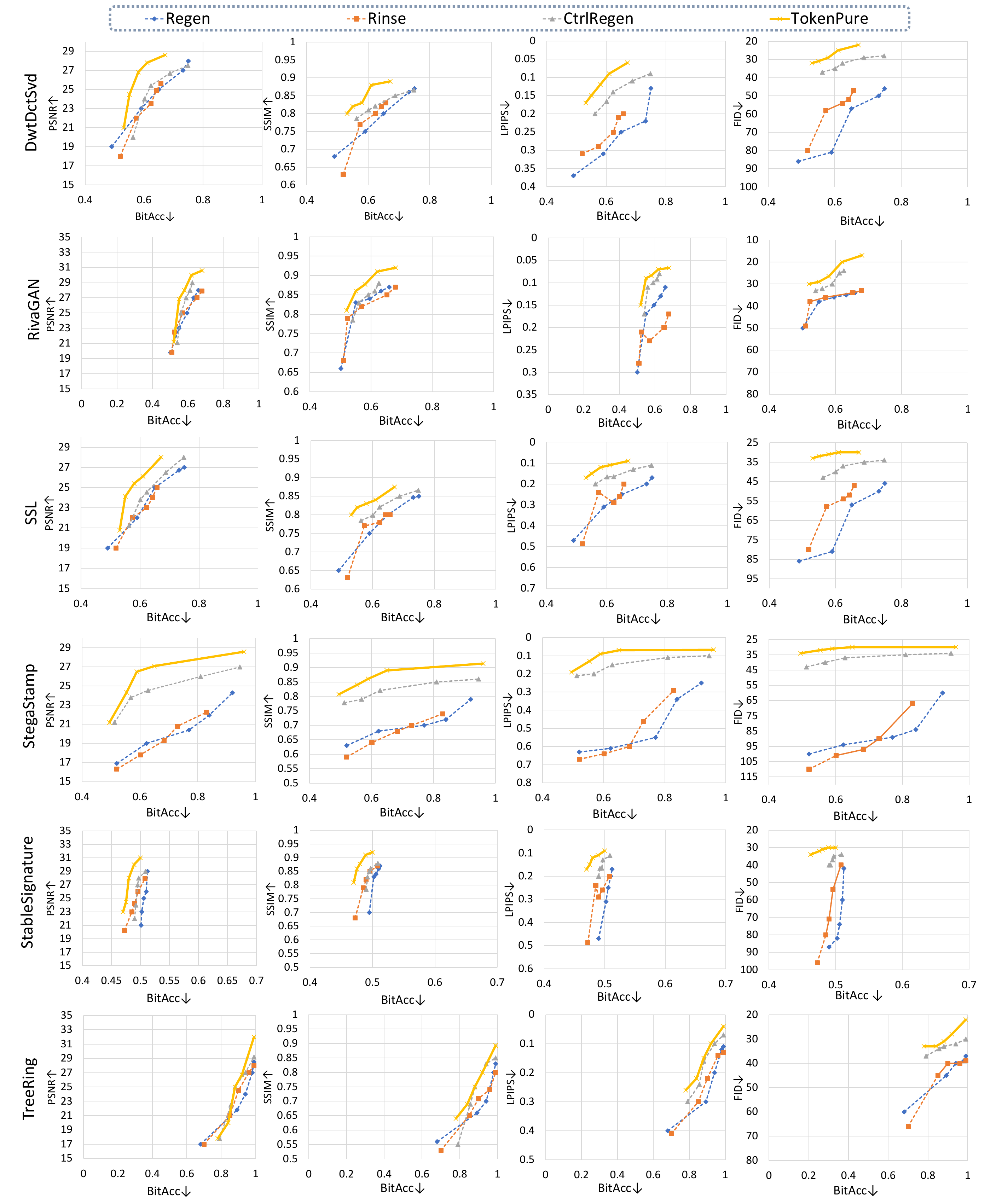}
  \caption{Performance of TokenPure compared to other methods with a varying number of noise strengths on different watermarks, respectively. We invert the LPIPS/FID scores to ensure that the top-left represents better performance across all figures. }
  \label{fig:line_chart_bitacc}
\end{figure*}

\begin{figure*}[h]
\centering
  \includegraphics[width=0.94\textwidth]{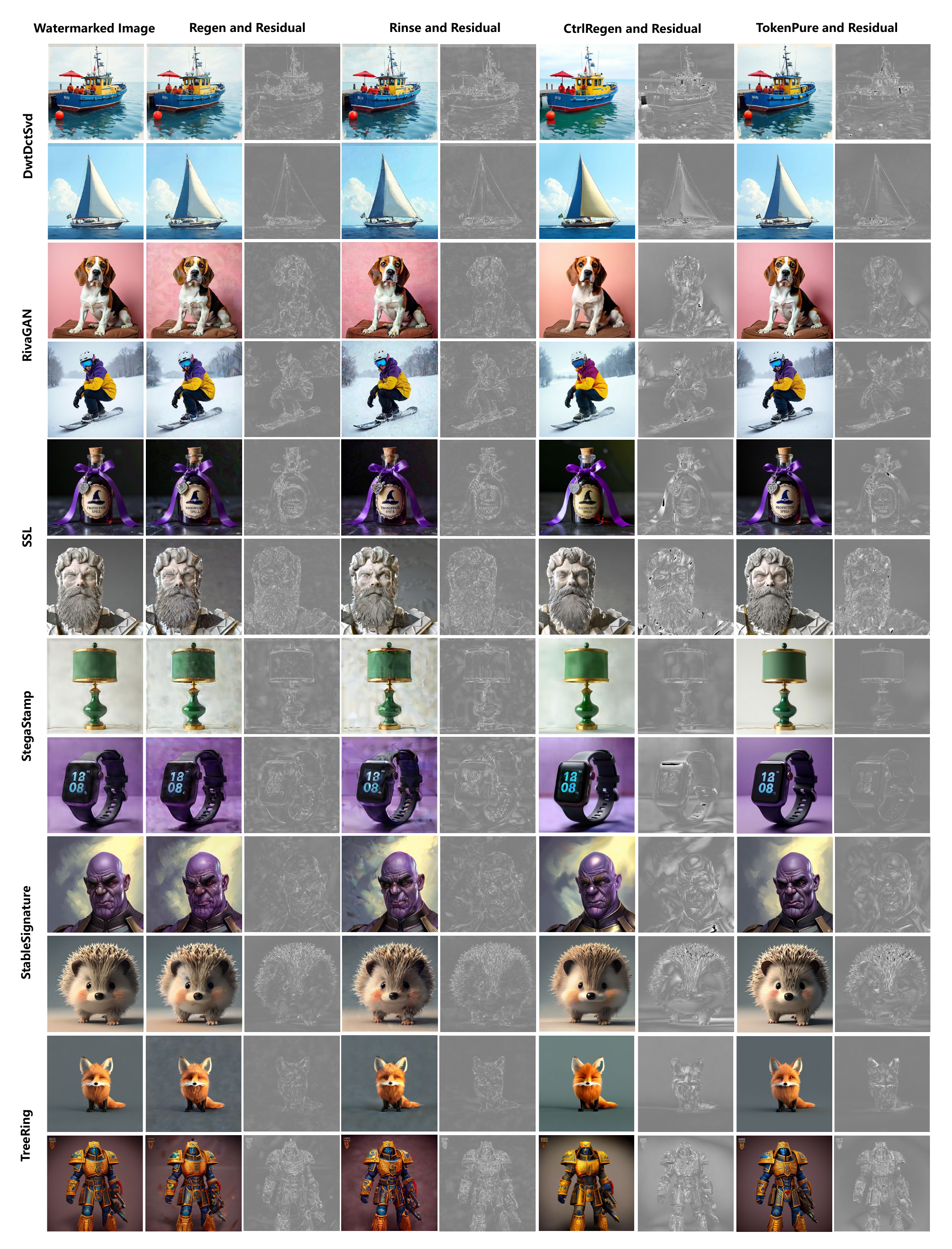}
      \caption{More qualitative comparison of different watermark removal attacks on different watermarking methods.} 
  \label{fig:visual_com_supp}
\end{figure*}

\textbf{Running efficiency} \textbf{Running efficiency} To assess TokenPure’s efficiency-performance trade-off, we analyze two key tables. Table \ref{tab:inferencetime} compares inference times across regeneration methods, while Table \ref{tab:diffenert_tokenpure} evaluates TokenPure’s performance with different denoising timesteps.
Table \ref{tab:inferencetime} reveals that TokenPure’s low-timestep variants (TokenPure-s4, TokenPure-s2) achieve substantial efficiency gains—with inference times reduced to 3.00s and 1.60s, respectively—compared to the full 25-timestep version (20.0s). Notably, these variants outpace competitors like Regen (0.60s, but with significant trade-offs in quality and removal efficacy) and CtrlRegen (2.56s, with semantic deviations).
Table \ref{tab:diffenert_tokenpure} further shows that fewer timesteps in TokenPure (e.g., s4, s2) introduce minor performance degradation in fine-grained detail reconstruction but retain core watermark removal capabilities. For instance, TokenPure-s4 maintains competitive BitAcc and TPR@1\%FPR across all watermark types, while even TokenPure-s2 outperforms the 25-timestep version in some metrics (e.g., BitAcc A for SSL). Although low-timestep variants lag slightly behind the full version in PSNR/SSIM, their efficiency improvement far outweighs the performance loss.
Compared to existing methods, TokenPure’s low-timestep variants, despite marginally lower PSNR/SSIM than Regen, lead in watermark removal effectiveness, visual consistency, and inference efficiency. Thus, TokenPure-s4/s2 strike an optimal balance between efficiency and performance while preserving watermark removal advantages.

\begin{table}[htbp]
  \centering
  \small
  \caption{Inference times of different regeneration methods.}
  \begin{tabular}{l c c}
    \hline
    {Attack Methods} & {Number of steps} & {Times of Inferring(s)} \\
    \hline
    Regen & 60 & 0.60 \\
    Rinse & 120 & 4.25 \\
    CtrlRegen & 50 & 2.56 \\
    TokenPure & 25 & 20.0 \\
    TokenPure-s4 & 4 & 3.00 \\
    TokenPure-s2 & 2 & 1.60 \\
    \hline
  \end{tabular}
  \label{tab:inferencetime}
\end{table}
\begin{table*}[h]
  \centering
  \caption{Evaluation results of TokenPure with different denoising timesteps. The best results are denoted as \textbf{bold}.}
  \label{tab:diffenert_tokenpure}
  \adjustbox{max width=\linewidth}{%
  \begin{tabular}{@{}c c c c c c c c c c c c c@{}}
    \toprule
    Watermarks & Attack Methods & BitAcc B $\uparrow$ & BitAcc A$\downarrow$ & TPR@1\%FPR B $\uparrow$ & TPR@1\%FPR A$\downarrow$ & PSNR (dB)$\uparrow$ & SSIM $\uparrow$ & LPIPS $\downarrow$ & DISTS $\downarrow$ & Q-Align $\uparrow$ & CLIP-IQA $\uparrow$ & FID $\downarrow$ \\

    \midrule
    \multirow{3}{*}{DwtDctSvd} 
    & TokenPure & 0.9669 & \textbf{0.5001} & 0.9423 & \textbf{0.0000} & \textbf{19.77} & \textbf{0.7935} & \textbf{0.1832} & \textbf{0.1217} & \textbf{4.345} & \textbf{0.7859} & \textbf{32.95} \\
    & TokenPure-s4 & 0.9669 & 0.5021 & 0.9423 & \textbf{0.0000} & 19.18 & 0.7892 & 0.1981 & 0.1413 & 3.835 & 0.7825 & 37.46 \\
    & TokenPure-s2 & 0.9669 & 0.5009 & 0.9423 & \textbf{0.0000} & 19.64 & 0.7901 & 0.1940 & 0.1581 & 3.510 & 0.7742 & 44.59 \\
    \midrule
    \multirow{3}{*}{RivaGAN} 
    & TokenPure & 1.000 & \textbf{0.5270} & 1.000 & \textbf{0.0053} & \textbf{21.12} & \textbf{0.8056} & \textbf{0.1563} & \textbf{0.1098} & \textbf{4.415} & 0.7728 & \textbf{30.66} \\
    & TokenPure-s4 & 1.000 & 0.5446 & 1.000 & 0.0091 & 20.72 & 0.8053 & 0.1740 & 0.1348 & 3.869 & \textbf{0.7769} & 37.00 \\
    & TokenPure-s2 & 1.000 & 0.5826 & 1.000 & 0.0084 & 20.43 & 0.7976 & 0.1953 & 0.1593 & 3.484 & 0.7753 & 44.56 \\
    \midrule
    \multirow{3}{*}{SSL} 
    & TokenPure & 1.000 & 0.5314 & 1.000 & 0.0040 & \textbf{20.77} & \textbf{0.7954} & \textbf{0.1770} & \textbf{0.1363} & \textbf{4.476} & \textbf{0.7733} & \textbf{33.05} \\
    & TokenPure-s4 & 1.000 & \textbf{0.5094} & 1.000 & \textbf{0.0000} & 20.27 & 0.7948 & 0.1931 & 0.1557 & 3.687 & 0.7728 & 40.53 \\
    & TokenPure-s2 & 1.000 & 0.5136 & 1.000 & 0.0060 & 19.94 & 0.7880 & 0.2084 & 0.1781 & 3.218 & 0.7669 & 47.32 \\
    \midrule
    \multirow{3}{*}{StegaStamp} 
    & TokenPure & 1.000 & \textbf{0.4947} & 1.000 & \textbf{0.0080} & \textbf{21.19} & \textbf{0.8074} & \textbf{0.1920} & \textbf{0.1347} & \textbf{4.197} & \textbf{0.7743} & \textbf{33.34} \\
    & TokenPure-s4 & 1.000 & 0.5113 & 1.000 & 0.0120 & 20.61 & 0.8031 & 0.1952 & 0.1465 & 3.720 & 0.7728 & 37.40 \\
    & TokenPure-s2 & 1.000 & 0.5025 & 1.000 & \textbf{0.0080} & 20.34 & 0.7964 & 0.2093 & 0.1665 & 3.330 & 0.7703 & 44.92 \\
    \midrule
    \multirow{3}{*}{StableSignature} 
    & TokenPure & 0.9848 & 0.4677 & 0.9960 & 0.0065 & \textbf{18.02} & \textbf{0.6154} & 0.2891 & 0.1806 & \textbf{3.899} & 0.7399 & \textbf{37.91} \\
    & TokenPure-s4 & 0.9848 & \textbf{0.4481} & 0.9960 & \textbf{0.0059} & 17.10 & 0.6054 & \textbf{0.3006} & 0.1865 & 3.603 & \textbf{0.7449} & 42.41 \\
    & TokenPure-s2 & 0.9848 & 0.4654 & 0.9960 & 0.0092 & 16.77 & 0.6128 & 0.3431 & \textbf{0.2150} & 2.871 & 0.7404 & 47.03 \\
    \midrule
    \multirow{3}{*}{TreeRing} 
    & TokenPure & 1.000 & \textbf{0.7725} & 1.000 & \textbf{0.1850} & \textbf{17.81} & \textbf{0.6461} & \textbf{0.2661} & \textbf{0.1545} & \textbf{4.045} & \textbf{0.7333} & \textbf{34.97} \\
    & TokenPure-s4 & 1.000 & 0.8017 & 1.000 & 0.2662 & 17.28 & 0.6450 & 0.2890 & 0.1789 & 3.303 & 0.7329 & 38.30 \\
    & TokenPure-s2 & 1.000 & 0.8320 & 1.000 & 0.3300 & 17.08 & 0.6423 & 0.3119 & 0.1992 & 2.906 & 0.7350 & 42.68 \\
    \bottomrule
  \end{tabular}
  }
\end{table*}

\textbf{More Quantitative results  }
Further comparisons across watermark types in another core metric, average Bit Accuracy (BitAcc) for watermark removal, confirm TokenPure’s advantages in removal effectiveness and image quality preservation. Figure \ref{fig:line_chart_bitacc} presents a comparative analysis of TokenPure against three alternative methods (Regen, Rinse, CtrlRegen) across six watermarking schemes (four low-perturbation ones: DwtDctSvd, RivaGAN, SSL, StableSignature; and two high-perturbation ones: StegaStamp, TreeRing). Unlike the metric TPR@1\%FPR, the variation of BitAcc with different noise intensities is relatively subtle. However, from the overall trend, our method achieves higher image quality metrics such as PSNR, LPIPS, and FID under the same BitAcc. Taking the watermark StegaStamp as a example, when BitAcc is reduced to a comparable level, TokenPure maintains a PSNR that is 2–3 dB higher and an FID that is 15–20 points lower than Regen and Rinse, indicating significantly better structural and perceptual consistency.

It is evident that our method TokenPure achieves effective watermark removal (low BitAcc) while retaining original image integrity, striking a superior balance between erasure efficacy and consistency, where competing methods either fail to reduce BitAcc sufficiently or suffer from severe image degradation, whereas TokenPure consistently delivers both robust watermark erasure and high-fidelity reconstruction.

\textbf{More Qualitative results  }
To further illustrate the superiority of TokenPure in visual consistency, we include additional visual comparisons in the supplementary materials (Figure \ref{fig:visual_com_supp}), where each watermarking case presents the attacked image and a residual map (the difference between the attacked and original unwatermarked images).

\end{document}